\documentclass{ieeetj}
\usepackage{tabularx}
\newcolumntype{C}[1]{>{\centering\arraybackslash\vspace{0pt}}m{#1}}
\newcolumntype{Y}{>{\centering\arraybackslash}X}
\newcolumntype{P}{>{\centering\arraybackslash}p}

\usepackage{bm}
\usepackage{multirow}

\usepackage{cite}
\usepackage{amsmath,amssymb,amsfonts}
\usepackage{algorithmic}
\usepackage{graphicx,color}
\usepackage{textcomp}
\usepackage{xcolor}
\usepackage{hyperref}
\hypersetup{hidelinks=true}
\usepackage{algorithm,algorithmic}
\usepackage{subfigure}
\def\BibTeX{{\rm B\kern-.05em{\sc i\kern-.025em b}\kern-.08em
    T\kern-.1667em\lower.7ex\hbox{E}\kern-.125emX}}
\AtBeginDocument{\definecolor{tmlcncolor}{cmyk}{0.93,0.59,0.15,0.02}\definecolor{NavyBlue}{RGB}{0,86,125}}

\def\OJlogo{\vspace{-4pt}$<$Society logo(s) and publication title will appear here.$>$}
\def\seclogo{\vspace{10pt}$<$Society logo(s) and publication title will appear here.$>$}

\def\authorrefmark#1{\ensuremath{^{\textbf{#1}}}}

\usepackage{xcolor}
\newcommand{\rev}[1]{#1}

\begin{document}
\receiveddate{XX Month, XXXX}
\reviseddate{XX Month, XXXX}
\accepteddate{XX Month, XXXX}
\publisheddate{XX Month, XXXX}
\currentdate{XX Month, XXXX}
\doiinfo{XXXX.2022.1234567}

\markboth{}{Author {et al.}}

\title{SAS: Semantic-aware Sampling for Generative Dataset Distillation}

\author{MINGZHUO LI \authorrefmark{1} (Graduate Student Member, IEEE), 
GUANG LI \authorrefmark{1} (Member, IEEE), \\
LINFENG YE \authorrefmark{2} (Graduate Student Member, IEEE), 
JIAFENG MAO \authorrefmark{3} (Member, IEEE), 
TAKAHIRO OGAWA \authorrefmark{1} (Senior Member, IEEE), 
KONSTANTINOS N. PLATANIOTIS \authorrefmark{2} (Fellow, IEEE),
AND MIKI HASEYAMA \authorrefmark{1} (Senior Member, IEEE)
}
\affil{Hokkaido University, N-14, W-9, Kita-Ku, Sapporo, 060-0814, Japan}
\affil{University of Toronto, 27 King's College Circle, Toronto, Ontario M5S 1A1, Canada}
\affil{The University of Tokyo, 7-3-1 Hongo, Bunkyo-ku, Tokyo, 113-8654, Japan}
\corresp{CORRESPONDING AUTHOR: MIKI HASEYAMA (email: mhaseyama@lmd.ist.hokudai.ac.jp).}


\authornote{This study was supported in part by JSPS KAKENHI Grant Numbers JP23K21676, JP24K02942, JP24K23849, and JP25K21218.}

\begin{abstract}
Deep neural networks have achieved impressive performance across a wide range of tasks, but this success often comes with substantial computational and storage costs due to large-scale training data. Dataset distillation addresses this challenge by constructing compact yet informative datasets that enable efficient model training while maintaining downstream performance. However, most existing approaches primarily emphasize matching data distributions or downstream training statistics, with limited attention to preserving high-level semantic information in the distilled data. In this work, we introduce a semantic-aware perspective for dataset distillation by leveraging Contrastive Language–Image Pretraining (CLIP) as a semantic prior for post-sampling. Our goal is to obtain distilled datasets that are not only compact but also semantically class-discriminative and diverse. To this end, we design three semantic scoring functions that quantify class relevance, inter-class separability, and intra-set diversity in a pretrained semantic space. Based on image pools generated by existing distillation methods, we further develop a two-stage strategy for effective sampling: the first stage filters semantically discriminative samples to form a reliable candidate set, and the second stage performs a dynamic diversity-aware selection to reduce redundancy while preserving semantic coverage. Extensive experiments across multiple datasets, image pools, and downstream models demonstrate consistent performance gains, highlighting the effectiveness of incorporating semantic information into dataset distillation.
\end{abstract}


\maketitle

\section{Introduction}
\label{sec:intro}
The rapid development of deep neural networks (DNNs) has driven significant progress across a wide range of fields \cite{zhan2025reviewSignal, Atoar2025reviewSignal}, including artificial intelligence and signal processing. By automatically learning expressive representations, DNNs effectively exploit large-scale data and achieve outstanding performance on complex tasks. However, this success heavily depends on massive training datasets, which introduce substantial computational and storage burdens \cite{strubell2019cost}. Large datasets require substantial memory resources, and training deep models on such data involves extensive computation time. For example, in image classification, DNN architectures such as DenseNet \cite{huang2017densNet} and ConvNext \cite{liu2022ConvNext} have demonstrated remarkable performance when trained on the ImageNet-1K dataset \cite{deng2015imageNet1k}, which exceeds 130 GB in size and often requires tens of hours of training. Similarly, training modern segmentation models such as SegFormer \cite{xie2021SegFormer} on the ADE20K \cite{zhou2019ADE20k} dataset typically takes several days. These high time and resource demands limit the accessibility and scalability of DNN-based solutions. Moreover, model development frequently involves repeated training runs for hyperparameter tuning, ablation studies, or adaptation to new data patterns, further amplifying the computational cost \cite{kaplan2020cost}.

\par

To address these challenges, researchers have investigated strategies for reducing the volume of training data while maintaining downstream performance. Among these approaches, dataset distillation \cite{wang2018datasetdistillation, li2022awesome} has attracted increasing attention. The goal of dataset distillation is to compress the original dataset into a compact, high-quality synthetic dataset such that models trained on the distilled dataset can achieve performance comparable to those trained on the original dataset. By substantially reducing the number of training samples, this paradigm enables faster model training and lower storage requirements. In addition, since the distilled data can serve as a substitute for the original data during downstream training, it offers potential benefits for privacy preservation \cite{li2020soft}. Beyond efficiency, dataset distillation also provides a valuable framework for analyzing data redundancy, studying model generalization, and supporting rapid prototyping and experimentation \cite{yu2023DDreview}. These properties make it an appealing research direction from both practical and theoretical perspectives.

\par

Since its introduction, dataset distillation has received substantial research attention, leading to rapid methodological advances \cite{liu2025DDreview}. Existing approaches can generally be divided into non-generative and generative categories, depending on how the distilled data are obtained. Non-generative methods, including early studies \cite{zhao2021DC, cazenavette2022MTT, li2023ddpp, li2025distMatch, li2024iadd} in this field, directly optimize a small set of synthetic samples by aligning training-induced statistics between the distilled and original datasets. In contrast, more recent works adopt generative formulations, where distilled data are produced through generative models, often combined with additional optimization objectives to better satisfy distillation goals \cite{wang2023DiM, gu2024minimax, li2024generative, li2025diversity, li2025diff}. The use of generative models naturally enables flexible data synthesis and improved scalability, which has contributed to their growing popularity in dataset distillation research. Among these generative approaches, diffusion models \cite{chen2024diffReview} have emerged as a dominant paradigm due to their high fidelity and strong generalization capability. Their success in modeling complex data distributions has further accelerated progress in generative dataset distillation.

\par

Despite recent advances, generative dataset distillation methods still exhibit notable limitations. Most existing approaches primarily focus on matching data distributions or downstream training statistics, which do not explicitly preserve high-level semantic information. As a result, the distilled samples may achieve statistical alignment while lacking clear semantic consistency and class-discriminative characteristics. Recent progress in vision–language models, particularly the Contrastive Language–Image Pretraining (CLIP) model \cite{radford2021clip}, provides a promising complementary semantic prior. Trained on large-scale image–text pairs, CLIP learns a task-agnostic embedding space where visual representations are aligned with semantic concepts and inter-class relationships are naturally encoded. Leveraging this property allows semantic information to be explicitly introduced into the distillation pipeline. By incorporating CLIP-based semantic guidance, the distillation can be driven not only by statistical criteria but also by semantic consistency, leading to subsets that are both semantically meaningful and class-discriminative. 

\par

In this paper, we propose Semantic-aware Sampling (SAS), a post-sampling framework that integrates high-level semantic information into generative dataset distillation. SAS operates on an image pool obtained by existing generative distillation methods and selects a compact subset that is semantically discriminative and diverse. Specifically, we leverage a pretrained CLIP model and introduce three complementary scoring functions that quantify class relevance, inter-class separability, and intra-set diversity in the semantic space. Based on these scores, we propose a two-stage strategy for effective sampling, including score filtering and diversity-aware selection. We evaluate SAS across multiple image pools, datasets, and downstream architectures. Experimental results demonstrate consistent performance gains over baseline methods, indicating that the integration of semantic information enhances the effectiveness of dataset distillation.

\par

The main contributions of this paper are summarized as follows:

\begin{itemize}
    \item \rev{We incorporate semantic information into the dataset distillation process by leveraging CLIP as a pretrained semantic prior, enabling the selection of semantically discriminative and diverse distilled datasets.}
    
    \item \rev{We propose three semantic scoring functions to quantify class relevance, inter-class separability, and intra-set diversity, and introduce a two-stage sampling strategy for effective implementation.}
    
    \item \rev{We conduct comprehensive experiments under various settings, including different image pools, datasets, and downstream architectures, demonstrating the effectiveness and generalization capability of the proposed method.}
\end{itemize}

\section{Related Work}
\subsection{Generative Models}
Generative models aim to learn the underlying data distribution and synthesize samples that resemble the original data \cite{richter2023genFrame}. By explicitly modeling the data-generating process, they are able to capture complex structures and variations within high-dimensional data. Early frameworks such as Variational Autoencoders (VAEs) \cite{kingma2014VAE} and Generative Adversarial Networks (GANs) \cite{goodfellow2014gan} provide two representative paradigms: latent-variable modeling and adversarial learning. With strong distribution modeling capabilities and high-quality synthetic samples, generative models have become an essential tool in current machine learning tasks.

\par

As for recent developments, diffusion models \cite{chen2024diffReview} define a forward diffusion process that gradually perturbs data with noise, and learn a corresponding reverse denoising process to recover clean samples, achieving the state-of-the-art (SOTA) performance. Based on the early framework, Denoising Diffusion Probabilistic Models (DDPMs) \cite{ho2020ddpm} reformulate the training objective for simplification. Subsequent advances include Stable Diffusion \cite{rombach2022stableDiffusion}, which operates in the latent space to reduce computational cost; Diffusion Transformer (DiT) \cite{Peebbles2023DiT}, which introduces transformer architecture to enhance global modeling capacity; and ControlNet \cite{zhang2023ContrlNet}, which enables conditional generation with additional branches. With enhanced model robustness and sample diversity, they have emerged as reliable foundational models for various applications, including image inpainting \cite{nichol2022imageInpaint}, speech enhancement \cite{Richter2024diffSpeechSignal}, and dataset optimization tasks such as dataset distillation \cite{su2024d4m, ye2025igds}.
\begin{figure}[t]
    \centering
    \subfigure[Non-generative methods.]{\label{fig:nGenDD}  \includegraphics[width=0.45\linewidth]{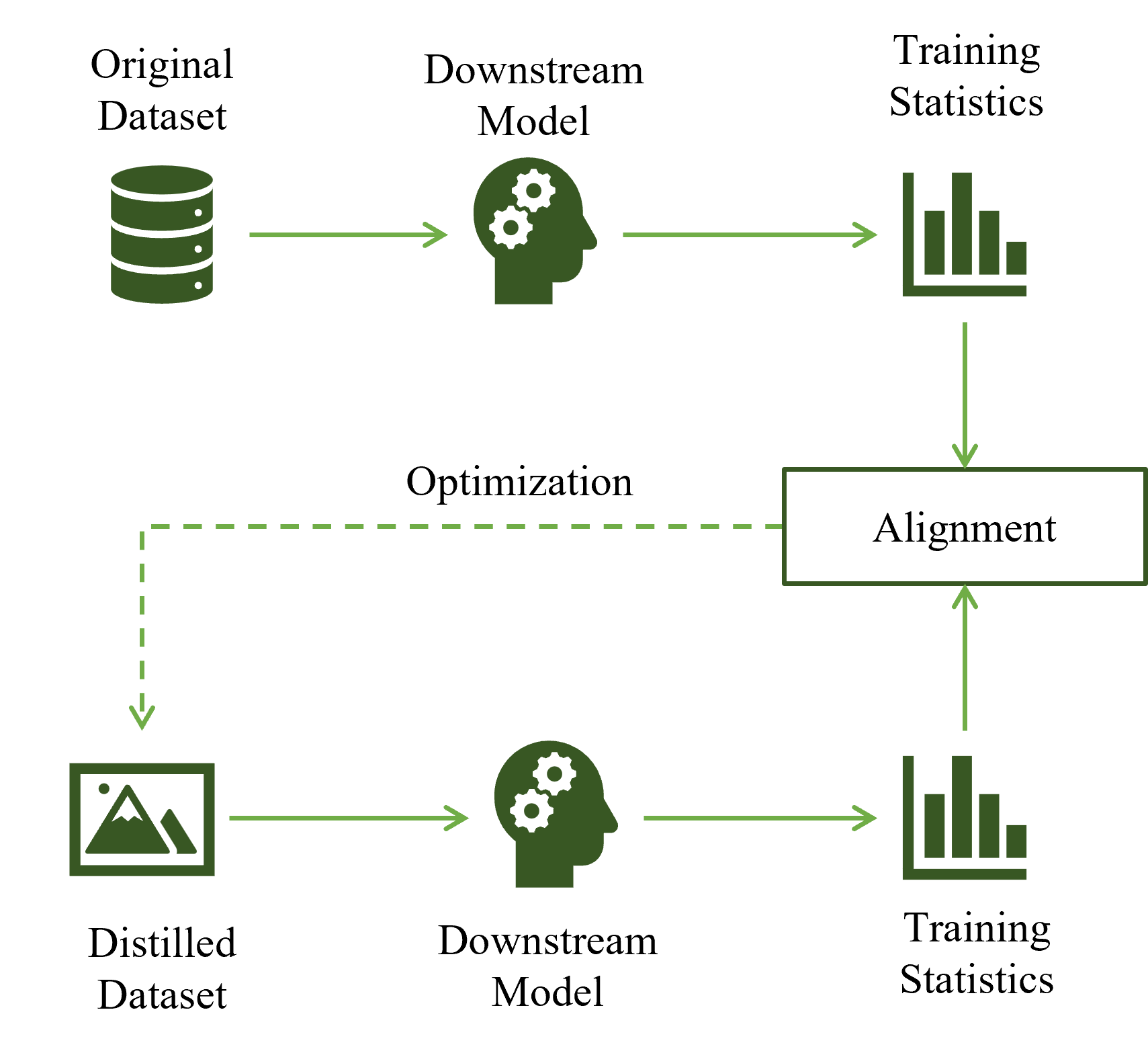}}
    \subfigure[Generative methods.]{\label{fig:GenDD} \includegraphics[width=0.45\linewidth]{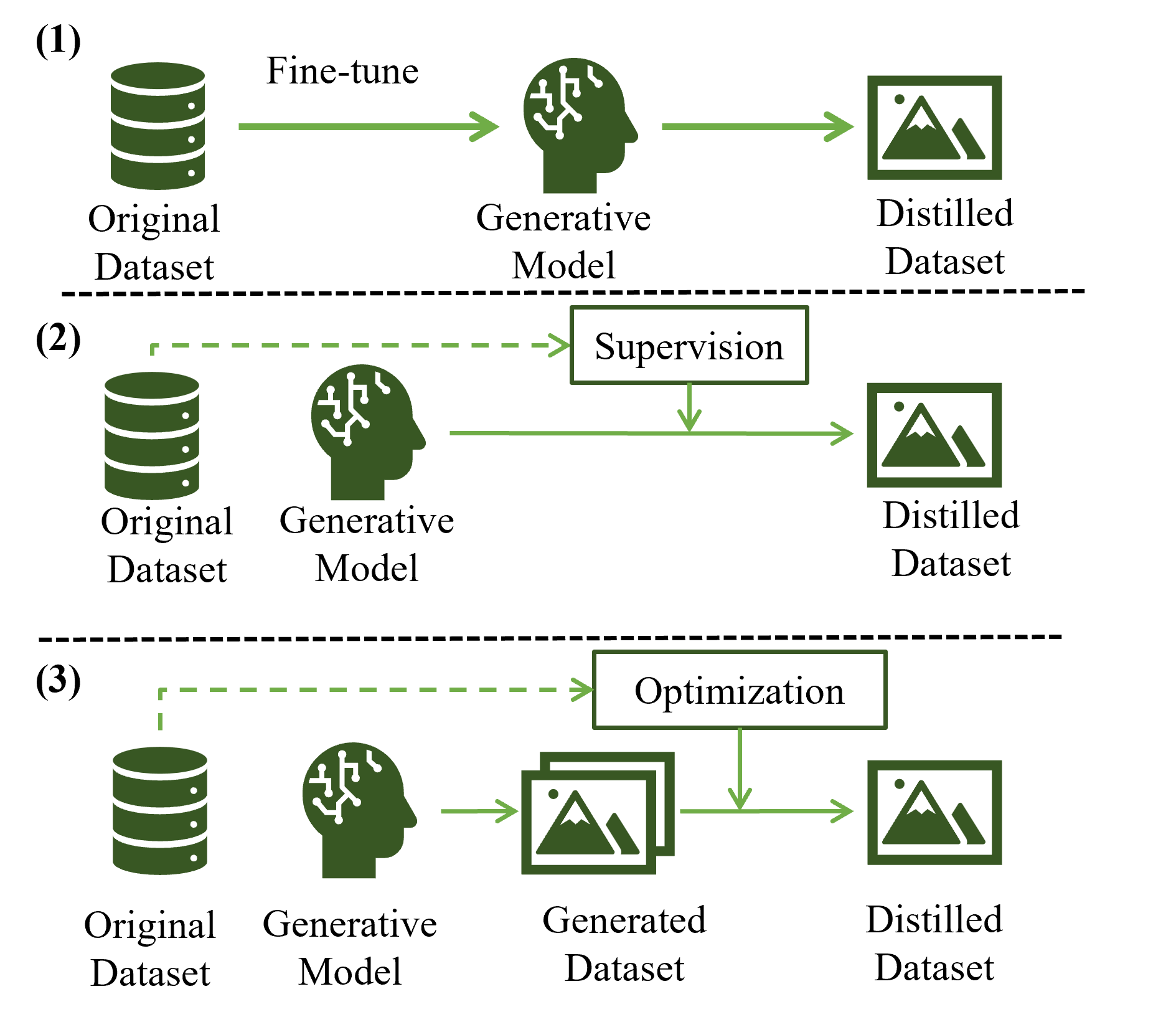}}
    \caption{Categories of current dataset distillation methods. Non-generative methods directly optimize distilled images by training-induced statistics. Generative methods rely on generative models to produce distilled datasets with three optimization paradigms: (1) fine-tuning of the generative model, (2) supervision on the generation pipeline, and (3) post-optimization on the distilled dataset.
    }
    \label{fig:DD}
\end{figure}

\subsection{Dataset Distillation}
Dataset distillation \cite{wang2018datasetdistillation} aims to compress a large-scale dataset into a compact set of representative samples while preserving training effectiveness. Models trained on the distilled dataset are expected to achieve comparable performance to those trained on the original dataset, with significantly reduced training time and computational cost. 
\rev{Different from traditional data selection methods, including pruning \cite{abbas2024pruning} and coreset selection \cite{moser2026coreset}, which mainly select or remove samples directly from the original dataset to reduce redundancy or computational cost, dataset distillation aims to obtain a compact synthetic set that preserves the knowledge and training dynamics of the original dataset. Recent methods include post-optimization on generated image pools, which may resemble traditional data selection methods but differ substantially in data distributions and optimization objectives.} As shown in Fig.~\ref{fig:DD}, depending on how the distilled data are obtained, existing methods can generally be divided into non-generative and generative categories.

\par
For non-generative methods, a small set of synthetic images is directly under optimization. Early methods commonly use training-induced statistics of the downstream models as optimization objectives. The pioneering work \cite{wang2018datasetdistillation} jointly trains a model on both real and distilled data and aligns their training losses. Subsequent works introduce alternative optimization objectives to reduce computational cost and improve alignment performance, including gradient dynamics\cite{zhao2021DC}, training trajectories\cite{cazenavette2022MTT}, data distribution\cite{li2025distMatch}, and kernel statistics\cite{chen2024kernal}. Recent decoupled methods reformulate dataset distillation by separating optimization, generation, and supervision. For instance, RDED \cite{sun2024RDED} utilizes the external realism score to supervise data synthesis. FADRM \cite{cui2025fadrm} decouples data generation and optimization by fusing intermediate optimized images with raw data patches.

\par
In contrast, generative methods use generative models to facilitate flexible data generation. To improve compression efficiency, existing methods have explored various paradigms, including fine-tuning of the generative model, supervision on the generation pipeline, and post-optimization on distilled data. For instance, DiM \cite{wang2023DiM} introduces a GAN model and uses logits for fine-tuning. MGD3 \cite{chan-santiago2025mgd3} uses class modes to guide the denoising process of DiT, and CaO$_2$ \cite{wang2025cao2} proposes a two-stage post-optimization to rectify objective inconsistency and conditional mismatch.

\par

Beyond these categories, recent works continue to explore new directions, such as dataset quantization \cite{li2025ddQuanti} and label distillation \cite{yu2025ddLabel}. Their applications also extend to broader domains, including architecture search \cite{gu2024DDcontinue}, privacy-preserving learning \cite{li2022compressed, zheng2025DDprivacy}, and recommendation system distillation \cite{zhang2025recommandDD}. 

\subsection{Semantic Representation}
CLIP learns a joint image–text embedding space through large-scale contrastive pretraining on both the text encoder and image encoder, enabling visual inputs and text descriptions to be aligned in a shared space. With strong semantic abstraction, CLIP exhibits remarkable transferability across various tasks, supporting zero-shot and few-shot recognition without requiring task-specific supervision \cite{sammani2024clipFrame}.

\par

CLIP has been widely adopted as a semantic guidance prior in image and audio generation, editing, and retrieval \cite{zhang2025clipImage, wu2025clipAudio}. By operating in a high-level semantic space, CLIP-based methods enable explicit control over semantic relevance and inter-class relationships, facilitating flexible conditioning via textual or conceptual constraints. In particular, CLIP has been extensively integrated with generative and diffusion models to guide sampling and optimization, demonstrating its effectiveness in controllable and semantically aligned generation \cite{bansal2023clipDiff}. These properties make CLIP a promising foundation for semantic-aware data distillation, where distilled data can preserve not only visual fidelity but also high-level semantic structures.

\section{Methodology}
This section is organized as follows. We first introduce the generative framework of DiT with Minimax used for constructing image pools, and the CLIP model used as a semantic encoder. We then describe the overall pipeline of SAS, including the formulation of the three semantic scoring functions and the proposed two-stage semantic-aware sampling strategy.

\begin{figure*}[t]
\centerline{\includegraphics[width=7in]{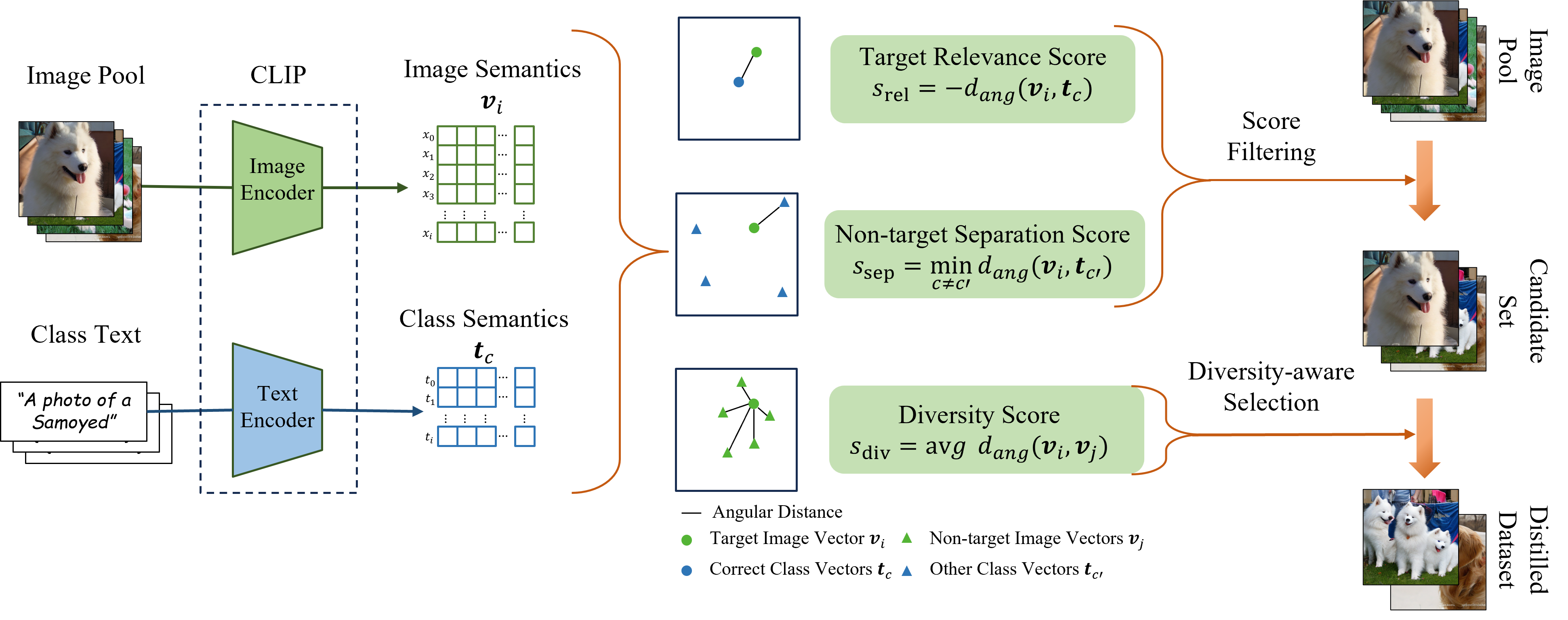}}
\caption{Calculation of semantic scores. The image pool is constructed using existing generative dataset distillation models and encoded with a pre-trained CLIP model, along with text prompts of the corresponding classes. Three semantic scores are calculated in the semantic space using angular distance.}
\label{fig_workflow}
\end{figure*}

\subsection{Preliminary}
\subsubsection{Generative Frameworks}
\label{2-1}
To construct a high-quality image pool for subsequent sampling, we adopt the DiT-based framework of Minimax \cite{gu2024minimax} as the generative baseline. Minimax introduces a min–max optimization objective for fine-tuning the DiT model in the dataset distillation setting. The DiT framework operates in the latent space and integrates transformer architectures with diffusion-based generation, providing a flexible and scalable generative backbone. The Minimax objective further encourages both representativeness with respect to the original data distribution and diversity among generated samples, making it well-suited for constructing image pools tailored for dataset distillation.

\par

Let the original dataset be denoted as $\mathcal{D} = \{{(\bm{x}_i, y_i)}\}_{i=1}^{N}$. Firstly, each image $\bm{x}_i$ is first encoded into a latent vector $\bm{z}_i$ using a pre-trained VAE encoder. The following formulations assume a batch size of one for notational simplicity. Given an image–label pair $(\bm{x}_0, y_0)$ with latent vector $\bm{z}_0$, Gaussian noise $\epsilon \in \mathcal{N}(\bm{0}, \bm{I})$ is gradually added to $\bm{z}_0$ over $t$ steps to obtain the noisy vector $\bm{z}_t$. This forward diffusion process is defined as follows:
\begin{equation}
    \bm{z}_t = \sqrt{\overline{\alpha}_t} ~ \bm{z}_0 + \sqrt{1 - \overline{\alpha}_t} ~ \epsilon, 
\end{equation}
where $\overline{\alpha}_t$ denotes the predefined variance schedule.

\par

The diffusion model $f_{\theta}$, parameterized by $\theta$, is trained to reverse this corruption process by predicting the added noise. Class information is incorporated via a class embedding vector $\bm{c}$ obtained from a class encoder applied to the label $y_0$. The standard diffusion objective minimizes the discrepancy between the predicted noise $f_{\theta}(\bm{z_t}, t, \bm{c})$ and the ground-truth noise $\epsilon$. This reverse denoising process is defined as follows: 
\begin{equation}
    \mathcal{L}_{\text{diffusion}} = {||f_{\theta}(\bm{z}_t, t, \bm{c}) - \epsilon||}_2^2.
\end{equation}
In the context of dataset distillation, diffusion models are typically initialized from pre-trained checkpoints. However, the conventional training objective alone is insufficient to provide explicit control for the distillation purpose.

\par

\rev{To address this limitation, Minimax introduces additional objectives to encourage both representativeness and diversity of the generated dataset to enhance distribution alignment. Specifically, two auxiliary memory sets consisting of latent vectors from real images and generated samples are maintained. Representativeness is quantified by the similarity between the generated samples and real images, yielding the representativeness loss that brings each generated sample closer to the least similar real sample. Conversely, diversity is measured by similarity among generated samples, yielding the diversity loss that pushes the current output away from the most similar generated sample. The overall training objective integrates the diffusion loss with the above Minimax losses, guiding the model toward producing datasets that are both representative of the original distribution and diverse within themselves \cite{gu2024minimax}.}

\vspace{-3mm}
\subsubsection{\rev{Semantic Embedding Space}}
\rev{Contrastive Language-Image Pretraining (CLIP) \cite{radford2021clip} is a vision-language foundation model trained on large-scale image-text pairs using contrastive learning. CLIP consists of an image encoder and a text encoder, which project images and text descriptions into a shared semantic embedding space. Through contrastive training, semantically related image-text pairs are mapped to nearby representations, while unrelated pairs are pushed farther apart.}

\par

\rev{Given an input image $\mathbf{x}$ and a text description $T$, CLIP extracts normalized image embedding $\mathbf{v}$ and text embedding $\mathbf{t}$ as follows: 
\begin{equation}
    \mathbf{v} = E_{\text{img}}(\mathbf{x}), \mathbf{t} = E_{\text{text}}(\mathbf{T}), 
\end{equation}
where $E_{\mathrm{img}}(\cdot)$ and $E_{\mathrm{text}}(\cdot)$ denote the image and text encoders, respectively. As the image and text are encoded into a shared embedding space, semantic similarity can be quantified using metrics such as cosine similarity or angular distance.}

\par

\rev{Due to the large-scale semantic pretraining, the CLIP embedding space captures rich semantic relationships across visual concepts. In this work, we utilize the pretrained CLIP embedding space for calculating semantic scores.}

\subsection{Semantic Sampling}
To approximate the data characteristics of the original dataset using a compact distilled dataset, we reformulate dataset distillation as a semantic coverage problem in a shared embedding space. The objective is to obtain a set of samples that preserves class-discriminative semantics while minimizing redundancy in the selected dataset. To achieve this, we employ CLIP as a unified semantic encoder, offering a pretrained embedding space where high-level semantic relationships between images and class prototypes can be explicitly quantified.

\par

We evaluate the semantic quality of each image from three complementary perspectives: class relevance, inter-class separability, and intra-set diversity. The first two criteria capture class-discriminative semantics, the third measures redundancy in the set. These properties are quantified by dedicated semantic scoring functions, which jointly guide the sampling process described in the following subsections.

\subsubsection{Semantic Scoring Functions}
The sampling is performed over an image pool $\mathbf{X}_\mathrm{IP}$ obtained by existing generative dataset distillation methods. For each image $\mathbf{x}_i \in \mathbf{X}_\mathrm{IP}$ with the corresponding class label $c$, we extract its CLIP-encoded $\ell_2$-normalized semantic feature $\mathbf{v}_i \in \mathbb{R}^d$. The semantic prototype of class $c$ is represented by the CLIP embedding of the text prompt ``A photo of a \{classname\}'', denoted as $\mathbf{t}_{c} \in \mathbb{R}^d$. Class relevance measures how well an image aligns with the semantic prototype of the corresponding class. 
\rev{Although cosine similarity is commonly used for semantic similarity measurement, it tends to compress numerical variations in high-similarity regions of the embedding space while exaggerating differences in lower-similarity regions. When similarity values calculated from different regions are included, directly using cosine similarity may underestimate meaningful semantic differences among high-similarity samples. Therefore, we adopt the following angular distance for similarity measurement, which better preserves relative semantic differences in the CLIP embedding space:}
\begin{equation}
\label{alg_ang}
d_{\mathrm{ang}}(\mathbf{v}_1, \mathbf{v}_2)
= \arccos \left( \mathbf{v}_1^\top \mathbf{v}_2 \right).
\end{equation}
where $\mathbf{v}_1$ and $\mathbf{v}_2$ are $\ell_2$-normalized CLIP features. In practice, we clamp the inner product to \rev{$[-1+\epsilon,\,1-\epsilon] \space (\epsilon=1e-6)$} to ensure numerical stability. With this angular distance, the defined scores are calculated based on image-text and image-image distances as shown in Fig.~\ref{fig_workflow}. \rev{Firstly, the target relevance score that measures the class relevance is defined as follows:
\begin{equation}
\label{alg_pos}
s_{\mathrm{rel}}(x_i, c)
= - d_{\mathrm{ang}}(\mathbf{v}_i, \mathbf{t}_c).
\end{equation}}\rev{Higher target relevance scores indicate that the images are more semantically aligned with their corresponding classes.}

\par

At the same time, to enhance discriminability, the inter-class separability is incorporated to measure how close an image is to its most confusing alternative class. \rev{It is defined as the minimum angular distance between the image and the semantic prototypes of all other classes, leading to the non-target separation score shown as follows:
\begin{equation}
\label{alg_sepe}
s_{\mathrm{sep}}(x_i, c)
= \min_{c' \in \mathcal{C},\, c' \neq c}
d_{\mathrm{ang}}(\mathbf{v}_i, \mathbf{t}_{c'}),
\end{equation}
}\rev{where $\mathcal{C}$ denotes the set of all classes in $\mathbf{X}_\mathrm{IP}$.} 
\rev{Higher non-target separation scores indicate weaker semantic similarity to non-target classes, which corresponds to clearer semantic boundaries between categories in the selected dataset.}
\begin{table*}
    \centering
    \footnotesize
    \renewcommand{\arraystretch}{1.5}
    
    \tabcolsep=5pt
    \caption{Comparison of downstream validation accuracy with various baseline methods using different downstream models. The experiments are conducted on ImageWoof. DiT/Minimax + SAS represents performing SAS on image pools generated by DiT/Minimax. \rev{The "Ratio" denotes the proportion of the distilled dataset size relative to the original dataset.} The highest accuracy values are highlighted in bold.}
    \begin{tabular}{c|c|ccc|cc|cc|cc|c}
        \hline
        IPC (Ratio) & Test Model & Random & K-Center \cite{sener2018kcenter} & IDC \cite{kim2022IDC} & DiT \cite{Peebbles2023DiT} & DiT + SAS & Minimax \cite{gu2024minimax} & Minimax + SAS & Full Dataset 
        \\ 
        \hline
        & ConvNet-6 
        & $24.3_{\pm 1.1}$ 
        & $19.4_{\pm 0.9}$ 
        & $33.3_{\pm 1.1}$ 
        & $31.4_{\pm 0.4}$ 
        & $30.5_{\pm 1.0}$ 
        & $34.1_{\pm 0.4}$ 
        & \bm{$36.6_{\pm 1.7}$}
        & $86.4_{\pm 0.2}$ 
        
        \\ 
        10 (0.8\%) 
        & ResNetAP-10 
        & $28.4_{\pm 0.7}$ 
        & $22.1_{\pm 0.1}$ 
        & $37.3_{\pm 0.4}$ 
        & $34.4_{\pm 0.6}$ 
        & $36.0_{\pm 0.2}$
        & $35.7_{\pm 0.3}$ 
        & \bm{$40.3_{\pm 0.5}$}
        & $87.5_{\pm 0.5}$ 
        \\
        
        & ResNet-18 
        & $27.7_{\pm 0.9}$ 
        & $21.1_{\pm 0.4}$ 
        & $36.9_{\pm 0.4}$ 
        & $33.9_{\pm 0.1}$ 
        & $34.1_{\pm 1.2}$
        & $35.3_{\pm 0.4}$ 
        & \bm{$39.0_{\pm 1.4}$} 
        & $89.3_{\pm 1.2}$ 
        \\ 
        \hline
        
        & ConvNet-6 
        & $29.1_{\pm 0.7}$ 
        & $21.5_{\pm 0.8}$ 
        & $35.5_{\pm 0.8}$ 
        & $36.9_{\pm 1.1}$ 
        & $39.2_{\pm 1.0}$ 
        & $36.9_{\pm 1.2}$ 
        & \bm{$38.4_{\pm 1.6}$} 
        & $86.4_{\pm 0.2}$ 
        \\ 
        
        20 (1.6\%) 
        & ResNetAP-10 
        & $32.7_{\pm 0.4}$ 
        & $25.1_{\pm 0.7}$ 
        & $42.0_{\pm 0.4}$ 
        & $41.1_{\pm 0.8}$ 
        & $42.5_{\pm 0.6}$ 
        & $43.3_{\pm 0.3}$ 
        & \bm{$45.0_{\pm 0.8}$}
        & $87.5_{\pm 0.5}$ 
        \\ 
        
        & ResNet-18 
        & $29.7_{\pm 0.5}$ 
        & $23.6_{\pm 0.3}$ 
        & $38.6_{\pm 0.2}$ 
        & $40.5_{\pm 0.5}$ 
        & $40.6_{\pm 0.7}$ 
        & $40.9_{\pm 0.6}$ 
        & \bm{$42.9_{\pm 0.8}$} 
        & $89.3_{\pm 1.2}$ 
        \\ 
        \hline
         
        & ConvNet-6 
        & $41.3_{\pm 0.6}$ 
        & $36.5_{\pm 1.0}$ 
        & $43.9_{\pm 1.2}$ 
        & $46.5_{\pm 0.8}$ 
        & $47.1_{\pm 0.7}$ 
        & $51.4_{\pm 0.4}$ 
        & \bm{$54.7_{\pm 0.6}$} 
        & $86.4_{\pm 0.2}$ 
        \\ 
        
        50 (3.8\%) 
        & ResNetAP-10 
        & $47.2_{\pm 1.3}$ 
        & $40.6_{\pm 0.4}$ 
        & $48.3_{\pm 1.0}$ 
        & $49.3_{\pm 0.2}$ 
        & $53.7_{\pm 0.4}$ 
        & $54.4_{\pm 0.6}$ 
        & \bm{$59.9_{\pm 0.8}$}
        & $87.5_{\pm 0.5}$ 
        \\ 
        
        & ResNet-18 
        & $47.9_{\pm 1.8}$ 
        & $39.6_{\pm 1.0}$ 
        & $48.3_{\pm 1.0}$ 
        & $50.1_{\pm 0.5}$ 
        & $51.7_{\pm 1.0}$ 
        & $53.9_{\pm 1.3}$ 
        & \bm{$57.9_{\pm 0.8}$} 
        & $89.3_{\pm 1.2}$ 
        \\
        \hline
    \end{tabular}
    \label{exp_main_gen}
\end{table*}

\par

While class-discriminative scores measure semantic alignment, using them alone may cause redundant selections. To mitigate this issue, we introduce an intra-set diversity score to measure the semantic redundancy of an image among the selected samples. \rev{It is defined as the average angular distance to all other samples of the same class in $\mathbf{X}_\mathrm{IP}$ as follows:}
\begin{equation}
\label{alg_div}
\rev{
s_{\mathrm{div}}(x_i, c)
= \frac{1}{|\mathbf{X}_\mathrm{IP}^c| - 1}
\sum_{\substack{x_j \in \mathbf{X}_\mathrm{IP}^c \\ j \neq i}}
d_{\mathrm{ang}}(\mathbf{v}_i, \mathbf{v}_j), 
}
\end{equation}
where $\mathbf{X}_\mathrm{IP}^c$ denotes all the images with the class label $c$ in $\mathbf{X}_\mathrm{IP}$.
\rev{Higher diversity scores indicate that the images capture complementary semantic variations within the set, thereby reducing redundancy caused by near-duplicate images with similar discriminative characteristics.}

\par

The proposed semantic scoring functions characterize class relevance, inter-class separability, and intra-set diversity from complementary perspectives, offering quantitative criteria for selecting semantically discriminative subsets with reduced redundancy. 

\subsubsection{Sampling Strategy}
With three quantified scores, a straightforward approach is to aggregate these scores into a single mixed score and rank all samples accordingly. However, these scores capture different aspects of semantic information. In particular, class-discriminative semantics evaluate the relationship between the image and classes through image–text alignment, whereas intra-set diversity is determined by image–image relationships within the selected set. Direct aggregation would require careful tuning of weighting coefficients and may obscure the distinct roles of these scores. Moreover, diversity should be evaluated with respect to the evolving selected set rather than the static image pool for accurate evaluation of redundancy.

\par

Based on this consideration, we adopt a two-stage adaptive sampling strategy. For the first stage, we construct a candidate set consisting of samples with strong class-discriminative semantics. Intuitively, a desirable sample should be close to its corresponding class prototype while maintaining distance from other class prototypes. We quantify this property using a margin score, combining the target relevance and non-target separation scores, which is defined as follows:
\begin{equation}
s_{\mathrm{margin}}(\bm{x}_i, c) = s_{\mathrm{rel}}(\bm{x}_i, c) + s_{\mathrm{sep}}(\bm{x}_i, c).
\end{equation}
Higher margin scores indicate that the images have stronger semantic discrimination. \rev{Then, the candidate set of $p \in (0, 1)$ times the size of the image pool is formed by selecting samples with high margin scores.} This step removes semantically ambiguous or noisy samples, providing a reliable basis for subsequent sampling.

\par
For the next stage, we perform diversity-aware selection from the candidate set to dynamically form the final distilled dataset for reduced redundancy. 
\rev{Starting from an empty set $S$, the current selected set $S_n$ is iteratively constructed by sequentially adding samples from the candidate set. When the size of $S_n$ exceeds the target number of images per class (IPC), the sample with the lowest diversity score is removed to maintain the desired set size. Instead of directly using global diversity scores precomputed with respect to the image pool $X_\mathrm{IP}$, we dynamically evaluate the diversity according to the current selection status by replacing $X_\mathrm{IP}$ in Equation \ref{alg_div} with $S_n$. In this way, the diversity relationship is adaptively updated throughout the sampling process. After all candidate samples have been examined, the final selected set $S'$ is used as the distilled dataset.} This greedy update procedure progressively improves semantic coverage while maintaining a fixed set size, discouraging the inclusion of near-duplicate samples.

\par

Overall, this two-stage strategy decouples discriminative semantic filtering from redundancy control, allowing each to be addressed independently without requiring manual weighting between different scores. Furthermore, diversity is evaluated dynamically based on the current selection status rather than static distances within the image pool, enabling adaptive and reliable subset construction.

\begin{table}
    \centering
    \footnotesize
    \renewcommand{\arraystretch}{1.5}
    \caption{Comparison of downstream validation accuracy with other baseline methods on various ImageNet subsets. Results are reported using the downstream architecture of ResNetAP-10. The highest accuracy values are highlighted in bold.}
    \begin{tabularx}{\linewidth}{c|P{0.4cm}|YYP{1.8cm}Y}
        \hline

        & IPC & Random & DiT \cite{Peebbles2023DiT} & Minimax \cite{gu2024minimax} & SAS \\

        \hline
        \multirow{3}*{\rotatebox{90}{ImageNette}} & 10 
        & $54.2_{\pm 1.6}$ 
        & $59.1_{\pm 0.7}$ 
        & $59.8_{\pm 0.3}$ 
        & \bm{$60.5_{\pm 0.2}$} 
        \\
        
        & 20 
        & $63.5_{\pm 0.5}$ 
        & $64.8_{\pm 1.2}$ 
        & $66.3_{\pm 0.4}$ 
        & \bm{$66.4_{\pm 0.3}$}
        \\
        
        & 50 
        & $76.1_{\pm 1.1}$ 
        & $73.3_{\pm 0.9}$ 
        & $75.2_{\pm 0.2}$ 
        & \bm{$76.1_{\pm 0.6}$}
        \\

        \hline
        \multirow{3}*{\rotatebox{90}{ImageIDC}} & 10  
        & $47.0_{\pm 0.6}$ 
        & $50.2_{\pm 0.7}$ 
        & $51.3_{\pm 0.4}$ 
        & \bm{$52.2_{\pm 0.6}$}
        \\
        
        & 20 
        & $51.3_{\pm 0.3}$ 
        & $56.4_{\pm 0.5}$ 
        & $57.5_{\pm 1.2}$ 
        & \bm{$57.7_{\pm 0.9}$} 
        \\
        
        & 50 
        & \bm{$64.9_{\pm 0.8}$} 
        & $62.1_{\pm 0.4}$ 
        & $63.4_{\pm 0.8}$
        & $64.0_{\pm 0.7}$
        \\
        \hline
    \end{tabularx}
    \label{exp:generalization}
\end{table}
\section{Experiments}
\label{sec_exp}
\subsection{Experimental Settings}
To evaluate the effectiveness of the proposed semantic-aware sampling (SAS), we conduct comprehensive experiments across various settings. For image pools, we consider datasets generated by DiT \cite{Peebbles2023DiT} and Minimax \cite{gu2024minimax}. For datasets, we use three commonly used 10-class subsets from the full ImageNet dataset: ImageWoof \cite{fastai2019imageNette}, ImageNette \cite{fastai2019imageNette}, and ImageIDC \cite{kim2022IDC}. ImageWoof contains ten dog breeds and is widely regarded as a challenging benchmark for image classification. In contrast, ImageNette consists of ten classes that are relatively easy for classification, while ImageIDC is constructed by randomly selecting ten classes from ImageNet, serving as a more common data pattern. For downstream evaluation, we employ three classification architectures: ConvNet-6 \cite{gidaris2018convNet}, ResNet-18 \cite{he2016resNetAP}, and ResNet-10 with average pooling (ResNetAP-10) \cite{he2016resNetAP}. All models are trained from scratch on the distilled datasets using a fixed learning rate of 0.01. Performance is measured by top-1 classification accuracy on the validation set. We repeat each experiment three times and report the mean accuracy along with the corresponding standard deviation.

\par

\begin{table}
    \centering
    \footnotesize
    \renewcommand{\arraystretch}{1.5}
    \caption{\rev{Comparison of downstream validation accuracy with baseline generative method (marked as Diffusion) on CIFAR-10. Results are reported using the downstream architecture of ResNetAP-10. The highest accuracy values are highlighted in bold.}}
    \begin{tabularx}{\linewidth}{c|P{0.4cm}|YYY}
        \hline

        & IPC & Random & Diffusion \cite{ketanmann2024software} & SAS \\

        \hline
        \multirow{3}*{\rotatebox{90}{CIFAR-10}} & 10 
        & $31.7_{\pm 0.5}$ 
        & $37.5_{\pm 0.2}$ 
        & \bm{$37.6_{\pm 0.8}$} 
        \\
        
        & 20 
        & $38.8_{\pm 0.8}$ 
        & $43.9_{\pm 0.2}$ 
        & \bm{$44.9_{\pm 0.3}$}
        \\
        
        & 50 
        & $48.1_{\pm 0.3}$ 
        & $50.8_{\pm 0.6}$ 
        & \bm{$51.4_{\pm 0.4}$}
        \\
        \hline
    \end{tabularx}
    \label{exp:cifar}
\end{table}
We first apply SAS to image pools generated by different baseline methods, validating on different downstream models. We then examine its generalizability by conducting experiments on other datasets. In addition, qualitative analyses are provided to offer intuitive insights into the sampling behavior. We further analyze the results on different hyperparameter choices and ablation settings to identify the appropriate configurations.

\par

For the DiT-based image pool, a DiT-XL-2 is used for generation. For the Minimax-based image pool, we follow the official implementation and parameter settings released by the authors of Minimax. Specifically, a pre-trained DiT-XL-2 equipped with DiffFit \cite{xie2023Difffit} is fine-tuned on the full subset for eight epochs using a batch size of 8. During training, input images are randomly shuffled, augmented, and resized to $256 \times 256$ before being encoded into the latent space via a VAE encoder. Optimization is performed using AdamW with a learning rate of $1 \times 10^{-3}$. In the diffusion stage, 50 denoising steps are applied. For SAS, a CLIP-ViT-B/32 is used as the semantic encoder. The image pool size is $4~ \times~ \mathrm{IPC}$, and the candidate set size is $0.5$ times that of the image pool.

\par

\subsection{Benchmark Results}
We evaluate the benchmark performance of SAS on the ImageWoof dataset under various IPC settings and across different downstream models. The image pools are obtained by generative methods, including DiT and Minimax. We compare the performance between SAS and the baseline methods along with other representative methods like random selection (Random), K-Center \cite{sener2018kcenter}, and IDC \cite{kim2022IDC}.

\par
\begin{figure}
        \centering
        \includegraphics[width=\linewidth]{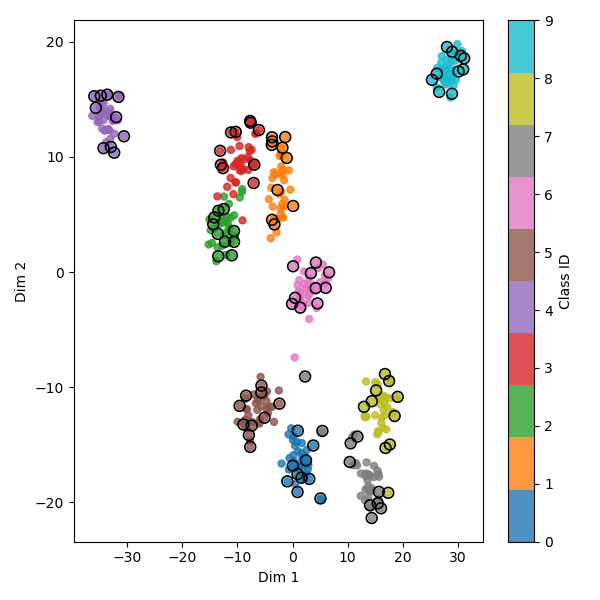}
        \caption{t-SNE analysis on the dataset sampled by SAS from the image pool. Overall, sampled datasets are distributed diversely inside the class cluster while maintaining relevance to the class center. The analysis is conducted on ImageWoof with IPC of 10.}
        \label{fig:tsne}
\end{figure}

As reported in Table~\ref{exp_main_gen}, SAS consistently achieves higher classification accuracy than baseline methods across most experimental configurations, with particularly notable improvements at higher IPC settings. These results demonstrate that incorporating semantic information into dataset distillation serves as an effective plug-in module that enhances the performance of the distilled datasets. Among the evaluated configurations, the combination of SAS and Minimax shows the best performance. Therefore, unless otherwise specified, Minimax is adopted as the default image pool generation method in subsequent experiments, being denoted directly as SAS for simplicity.

\par

\rev{We further evaluate the generalization ability of SAS by extending the experiments to additional ImageNet-based subsets as well as CIFAR-10 \cite{krizhevsky2009cifar}, which differs substantially in image resolution, semantic granularity, and visual distribution. Since the DiT model used in the main experiments is pretrained on ImageNet categories that are incompatible with CIFAR-10 labels, we adopt an open-source class-conditioned diffusion training framework \cite{ketanmann2024software} based on DDPM \cite{ho2020ddpm} to construct the CIFAR-10 image pools. As shown in Table~\ref{exp:generalization}, SAS exhibits similar performance trends on ImageNette and ImageIDC to those observed on ImageWoof. Furthermore, the results in Table~\ref{exp:cifar} demonstrate that SAS also consistently improves the performance of diffusion-based datasets on CIFAR-10. Overall, SAS achieves consistent improvements across different datasets and visual domains, indicating good generalization capability of the proposed framework.}

\subsection{Visualization}
To offer a qualitative understanding, we visualize the distribution of the selected images with respect to the full image pool. Specifically, we project CLIP features of all the classes into a two-dimensional space using t-SNE and highlight the locations of the selected images. The analysis is conducted on the ImageWoof dataset with an IPC of 10. As shown in Fig.~\ref{fig:tsne}, the samples in the selected dataset distribute diversely within the class, indicating the effectiveness of intra-set diversity. Although several samples appear to be located far from the main cluster of the corresponding class, most selected samples remain close to the class center. Such outlying points can be attributed to the intrinsic properties of t-SNE, which prioritizes preserving local neighborhood relationships, making the results relative to the image pool instead of the real global distribution. In addition, the discrepancy between the angular distance used during sampling and the distance metric employed by t-SNE may also contribute to these deviations.

\par

\begin{figure}
        \centering
        \includegraphics[width=\linewidth]{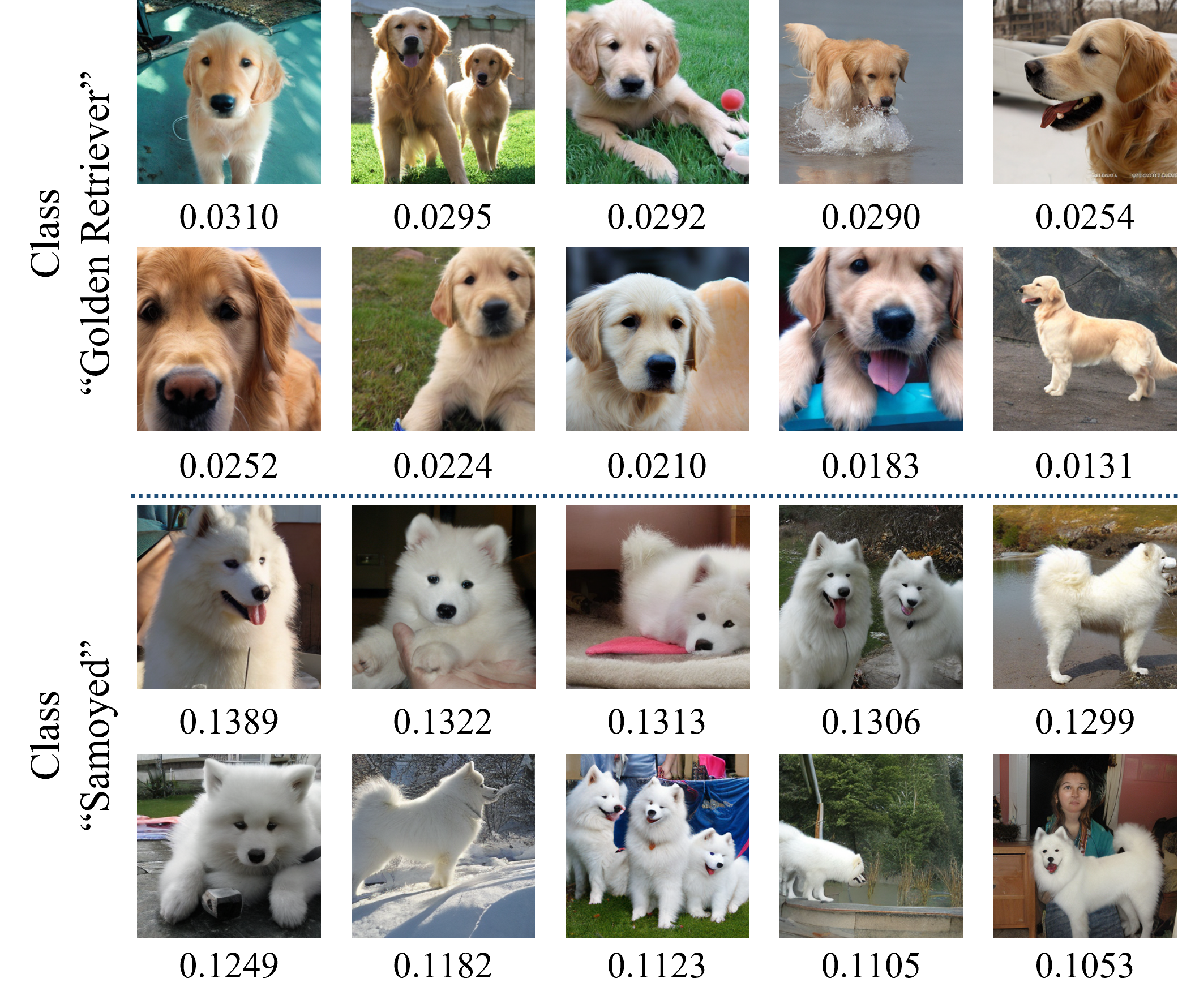}
        \caption{Representative images selected by SAS with corresponding margin scores. The dataset preserves class-discriminative images with high margin scores and maintains set diversity.}
        \label{fig_visual}
\end{figure}

We also show representative sampled images in Fig.~\ref{fig_visual} along with their corresponding margin scores, using two classes from ImageWoof as examples. The relative value of the margin score evaluates the class discriminative with respect to the corresponding class.
Results show that the selected dataset preserves mostly images with high margin scores, and several low-score images for diversity. Visually, the selected images preserve obvious class-discriminative visual characteristics while exhibiting diverse details within each class. The difference in absolute margin score values also provides potential for statistical analysis of class-specific data patterns in the semantic embedding space. 

\par

\begin{table}
    \centering
    \footnotesize
    \renewcommand{\arraystretch}{1.5}
    \caption{Comparison of downstream validation accuracy across different image pool sizes. The experiments are conducted on ImageWoof using ResNetAP-10. The highest accuracy values are highlighted in bold.}
    \centering
    \begin{tabularx}{\linewidth}{Y|YYY}
        \hline
        Size &  IPC=10 & IPC=20 & IPC=50
        \\
        \hline
        3 $\times$ IPC 
        & $38.1_{\pm 0.7}$ 
        & \bm{$45.4_{\pm 0.3}$}
        & $55.1_{\pm 0.4}$
        \\
        4 $\times$ IPC 
        & \bm{$40.3_{\pm 0.5}$} 
        & $45.0_{\pm 0.8}$
        & \bm{$59.9_{\pm 0.8}$} 
        \\
        
        5 $\times$ IPC 
        & $35.0_{\pm 1.2}$ 
        & $41.3_{\pm 0.2}$ 
        & $58.2_{\pm 0.7}$
        \\
        
        \hline
    \end{tabularx}
    \label{exp_ip_size}
\end{table}

\subsection{Size of Image Pool}
\label{sec_ip}
We further analyze the impact of the image pool size on the sampling results. Specifically, we generate image pools with different sizes of $n~ \times~ \mathrm{IPC}$, and examine how varying $n$ affects the quality of the distilled dataset. As shown in Table~\ref{exp_ip_size}, when the image pool is relatively small, the candidate set may not contain sufficiently diverse high-quality samples. In contrast, an excessively large image pool may introduce semantically ambiguous samples that receive high intra-set diversity scores. Therefore, an intermediate pool size provides a better trade-off between semantic discrimination and diversity. For the trade-off between performance and computational cost, the image pool size of $4~ \times~ \mathrm{IPC}$ is used as the default value. 

\subsection{\rev{Ratio of Candidate Set}}
\begin{figure}
        \centering
        \includegraphics[width=\linewidth]{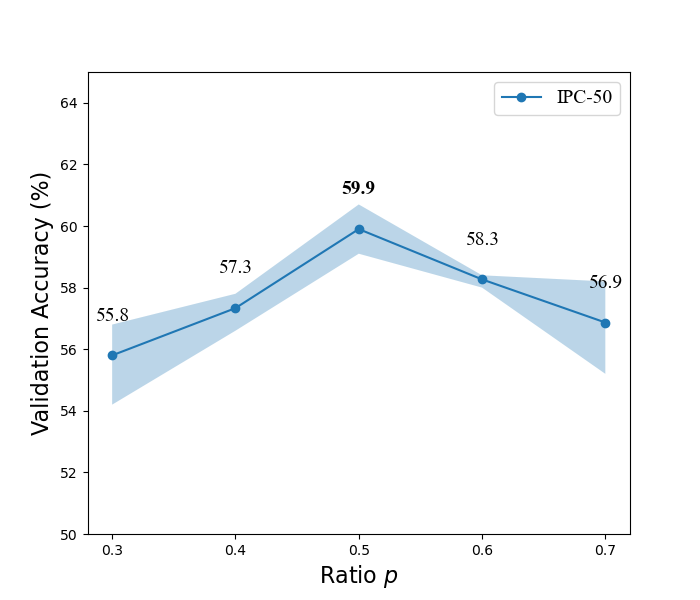}
        \caption{\rev{Comparison between different ratios of the candidate set. The experiments are conducted on ImageWoof using ResNetAP-10. The highest accuracy value is highlighted in bold.}}
        \label{exp:candidate}
\end{figure}

\rev{
We further investigate the influence of the candidate set size, which is defined as the retained ratio of the image pool ($p \times |\mathbf{X}_\mathrm{IP}|$) after the first-stage filtering. Specifically, using the same image pool, we construct candidate sets with different retained ratios and compare the downstream classification accuracy of the resulting distilled datasets. Considering that in small image pools, changes in ratio result in very small changes in image numbers, the experiments are conducted only under the IPC of 50. As the results shown in Fig.~\ref{exp:candidate}, the candidate set ratio significantly affects downstream performance. Similarly to the discussion in Section~\ref{sec_exp}.\ref{sec_ip}, the candidate pool potentially removes semantically diverse samples in small retained ratios and retains ambiguous samples in large retained ratios, both reducing the discriminative quality of the final distilled dataset. These observations suggest that the candidate set ratio helps to control the balance between semantic discrimination and diversity preservation during the two-stage sampling process. The retained ratio of $0.5$ is used as the default value.
}

\subsection{Sampling Strategy}
\label{sec:exp_samp}
We compare the proposed two-stage sampling strategy with the approach that directly ranks and selects samples using a single mixed score. To ensure comparability of class-discriminative semantics and intra-set diversity, we normalize the diversity score to align the distributions between image-text distance and image-image distance. Moreover, a weighting coefficient $\lambda$ is included to balance the effects of the two parts, leading to the mixed score as follows:
\begin{equation}
s_{\mathrm{mixed}}(\bm{x}_i, c) = s_{\mathrm{margin}}(\bm{x}_i, c) + \lambda ~ s^\mathrm{norm}_{\mathrm{div}}(\bm{x}_i, c).
\end{equation}
In this approach, images in $\mathbf{X}_\mathrm{IP}$ are ranked with the mixed score, and the highest $IPC$ samples are selected as the distilled dataset.

\par

As shown in Table~\ref{exp:samp_stra}, we conduct experiments using mixed scores with different $\lambda$ and compare the results with SAS. While the mixed-score approach provides a straightforward solution, it consistently underperforms SAS across all evaluated settings. This performance gap suggests that supplementary measures, including normalization and weighting coefficients, are insufficient to balance the class-discriminative semantics and intra-set diversity, supporting the effectiveness of diversity-aware selection in SAS. These results also indicate that treating diversity as a dynamic value relative to current status, rather than a static value calculated on the image pool, leads to more effective sampling. 

\begin{table}
    \centering
    \footnotesize
    \renewcommand{\arraystretch}{1.5}
    \caption{Comparison of downstream validation accuracy between SAS and sampling using mixed score. The experiments are conducted on ImageWoof using ResNetAP-10. The highest accuracy values are highlighted in bold.}
    \centering
    \begin{tabularx}{\linewidth}{c|YYY}
        \hline
        Method ($\lambda$) &  IPC = 10 & IPC = 20 & IPC = 50
        \\
        \hline
        Mixed Score (0.05) 
        & $34.3_{\pm 0.4}$ 
        & $38.9_{\pm 0.6}$ 
        & $57.1_{\pm 0.7}$ 
        \\
        
        Mixed Score (0.10) 
        & $33.6_{\pm 0.9}$ 
        & $40.4_{\pm 1.1}$ 
        & $56.5_{\pm 1.2}$
        \\
        
        Mixed Score (0.20) 
        & $33.6_{\pm 0.8}$ 
        & $41.5_{\pm 0.9}$ 
        & $54.6_{\pm 0.5}$
        \\
        
        SAS 
        & \bm{$40.3_{\pm 0.5}$} 
        & \bm{$45.0_{\pm 0.8}$} 
        & \bm{$59.9_{\pm 0.8}$}
        \\
        
        \hline
    \end{tabularx}
    \label{exp:samp_stra}
\end{table}

\subsection{Ablation Study}
We conduct an ablation study to analyze the contribution of each component in SAS, including the target relevance score, non-target separation score, and diversity-aware selection. Removing all components corresponds to random selection.

\par

Table~\ref{exp:ablation} reports the results of the ablation experiments. Without diversity-aware selection, the target relevance score leads to degraded performance that even underperforms random selection, possibly due to the selection of too many near-duplicate samples. Although the diversity-aware selection alone shows improved performance, it may include out-of-distribution samples with low class-discriminative semantics. SAS combines the two parts by filtering highly discriminative samples and constructing a diverse selected set. The ablation results demonstrate that all three components play complementary roles in SAS, and removing any one of them degrades the sampling performance.
\begin{table}
    \centering
    \footnotesize
    \renewcommand{\arraystretch}{1.5}
    \caption{Ablation Study on components of SAS. The experiments are conducted on ImageWoof using ResNetAP-10. The highest accuracy values are highlighted in bold.}
    \centering
    \begin{tabularx}{\linewidth}{ccc|YYY}
        \hline
        $s_\mathrm{rel}$ & $s_\mathrm{sep}$ & $s_\mathrm{div}$ &  IPC = 10 & IPC = 20 & IPC = 50
        \\
        \hline
        $\times$ & $\times$ & $\times$ 
        & $35.4_{\pm 0.6}$ 
        & $44.0_{\pm 1.3}$ 
        & $56.9_{\pm 0.5}$
        \\
        
        $\checkmark$ & $\times$ & $\times$ 
        & $34.3_{\pm 0.8}$ 
        & $41.2_{\pm 0.7}$ 
        & $55.3_{\pm 0.5}$ 
        \\
        
        $\times$ & $\checkmark$ & $\times$ 
        & $36.6_{\pm 0.3}$ 
        & $44.3_{\pm 1.3}$ 
        & $58.3_{\pm 0.2}$
        \\
        
        $\times$ & $\times$ & $\checkmark$ 
        & $37.5_{\pm 0.5}$ 
        & $44.8_{\pm 0.3}$ 
        & $59.0_{\pm 0.3}$
        \\
        
        $\checkmark$ & $\checkmark$ & $\times$ 
        & $34.9_{\pm 0.3}$ 
        & $42.5_{\pm 0.2}$ 
        & $55.8_{\pm 1.1}$
        \\
        
        $\checkmark$ & $\checkmark$ & $\checkmark$ 
        & \bm{$40.3_{\pm 0.5}$} 
        & \bm{$45.0_{\pm 0.8}$} 
        & \bm{$59.9_{\pm 0.8}$}
        \\
        \hline
    \end{tabularx}
    \label{exp:ablation}
\end{table}

\subsection{\rev{Computational Overhead}}
\rev{As a post-sampling strategy, SAS should not introduce substantial computational overhead to the original dataset distillation framework. Compared with the generation process, the memory overhead of CLIP-based sampling is relatively small, since CLIP contains fewer parameters (approximately 88 million) than DiT (approximately 676 million) and requires only a single forward pass without iterative denoising. Furthermore, CLIP and DiT are not executed simultaneously, resulting in no additional peak memory consumption.
}

\par

\rev{We further analyze the runtime overhead of SAS using an ImageWoof image pool containing 2,000 images. In our experimental setting using a single NVIDIA RTX A6000, SAS requires approximately 23 seconds (18 seconds for scoring and 5 seconds for sampling). Although this runtime is longer than that of K-Center \cite{sener2018kcenter} using the ResNet-50 as feature extractor (approximately 9 seconds) , the additional cost remains negligible compared with the image generation process (2,100 seconds). These results indicate that SAS introduces modest computational overhead while providing consistent improvements in downstream performance.}

\section{Conclusion}
\rev{In this paper, we propose the Semantic-aware Sampling (SAS), which leverages CLIP to investigate dataset distillation from a semantic perspective. By filtering and dynamically selecting samples in the semantic space, SAS obtains distilled datasets that are discriminative to the target class and are diversely distributed within the distilled dataset. We propose three semantic scoring functions to evaluate the class relevance, inter-class separability, and intra-set diversity, and introduce a two-stage sampling strategy for effective implementation. Extensive experiments across various settings and qualitative visualizations demonstrate the effectiveness of SAS. Ablation studies further confirm the complementary roles of each scoring function and the effectiveness of the proposed two-stage semantic-aware sampling strategy.
}

\par

\rev{
The effectiveness of SAS highlights the potential of incorporating high-level semantic information into generative dataset distillation and promotes further exploration of semantic-driven frameworks. Nevertheless, the proposed method relies on semantic priors extracted from pretrained CLIP embeddings, whose effectiveness may vary across out-of-distribution domains. In domains where CLIP lacks reliable semantic alignment, such as specialized medical images or remote-sensing imagery, the proposed scores may become less informative unless domain-adaptive semantic encoders are employed. Therefore, future work includes extending SAS to more recent generative frameworks, broader datasets, and visual domains, and evaluating its applicability to more diverse downstream tasks beyond image classification. We also consider investigating alternative semantic foundation models and domain-adaptive semantic encoders to further improve the robustness and generality of the semantic-aware dataset distillation framework.
}

\section*{Acknowledgments}
This study was supported in part by JSPS KAKENHI Grant Numbers JP23K21676, JP24K02942, JP24K23849, and JP25K21218.

\bibliographystyle{IEEEtran}
\bibliography{ref}

\end{document}